\documentclass[runningheads]{llncs}
\usepackage[T1]{fontenc}
\usepackage{amsmath,amssymb}
\usepackage{graphicx}
\usepackage{tikz}
\usepackage{algorithm}
\usepackage{algpseudocode}
\usepackage{multirow}
\usepackage{multicol}
\usepackage[numbers]{natbib}
\usepackage{booktabs}    % in preamble – nicer horizontal rules
\tikzset{
  observed/.style={circle, draw, fill=black, inner sep=1.5pt,
                   font=\scriptsize, text=white},
  latent/.style  ={circle, draw, dashed, fill=white,
                   inner sep=1.5pt, font=\scriptsize},
}

\usetikzlibrary{arrows.meta, positioning, graphs, quotes}

\newcommand{\val}{\mathrm{val}}

\newcommand{\graph}[1]{\mathcal{#1}}
\newcommand{\pa}[1]{\mathrm{Pa}(#1)}

\newcommand{\Do}[1]{\mathrm{do}(#1)}

\begin{document}
% arXiv: removed graphdrawing library (LuaTeX-only)
% algoritmo Sugiyama (hierárquico)

%\title{Root Causal Analysis with 
%Partially Identifiable Quasi-Markovian Models}

%\title{Probabilities of Causation for Root Cause Analysis with Limited Observability}

\title{Probabilities of Causation and Root Cause Analysis with Quasi-Markovian Models}
\titlerunning{Probabilities of Causation and Root Causal Analysis}

%\title{Limited Observability with Root Cause Analysis using/based on Probabilities of Causation}

\author{Eduardo Rocha Laurentino\inst{1}\inst{3} \and
Fabio Gagliardi Cozman\inst{2} \and
Denis Deratani Mau\'a\inst{2} \and
Daniel Angelo Esteves Lawand\inst{2} \and
Davi Goncalves Bezerra Coelho\inst{3}  \and
Lucas Martins Marques\inst{2}}

\institute{$^1$Instituto de Ciência e Tecnologia Itaú (ICTi) \\ $^2$Escola Politécnica, Universidade de S\~ao Paulo (EP-USP) \\ $^3$Instituto de Matemática e Estatística, Universidade de S\~ao Paulo (IME-USP)}

\authorrunning{Laurentino et al.}
\maketitle   

\begin{abstract}
Probabilities of causation provide principled ways to assess causal relationships but face computational challenges due to partial identifiability and latent confounding. This paper introduces both algorithmic simplifications, significantly reducing the computational complexity of calculating tighter bounds for these probabilities, and a novel methodological framework for Root Cause Analysis that systematically employs these causal metrics to rank entire causal paths.

\end{abstract}

\section{Introduction}

Causal reasoning underpins intelligent decision-making and effective problem-solving across diverse   domains. Several frameworks for causal inference have emerged, with Pearl's Structural Causal Model (SCM)
being particularly influential due to its rigorous algorithmic foundation and applicability \cite{primer}. Pearl's approach has seen successful adoption in various contexts, including medicine \cite{Richens2020}, manufacturing \cite{Oliveira2022}, and reliability engineering.

In particular, {\em probabilities of causation}, such as Probability of Necessity (PN), Probability of Sufficiency (PS) and Probability of Necessity and Sufficiency (PNS), have demonstrated value in multiple fields, such as medical diagnostics, explainable artificial intelligence (XAI), climate change analysis and representation learning. Recent works highlight their effectiveness: Watson et al. (2022) developed local explanations explicitly utilizing necessity and sufficiency; Cai et al. (2022) optimized GNN explanations by maximizing PNS; and Yang et al. (2023) employed a PNS-based risk metric for robust representation learning. These examples illustrate the practical relevance and theoretical robustness of these causal probabilities.

However, the calculation of probabilities of causation poses significant computational challenges. In particular, probabilities of causation  often suffer from partial identifiability issues. Latent confounding and non-monotonic relationships prevent exact identification, forcing reliance on computationally intensive optimization methods that yield intervals rather than precise values. These computations involve multilinear programming subject to linear constraints, representing a substantial barrier for practical applications.           

Addressing this complexity, we first provide theoretical advancements that significantly simplify these multilinear programs. By extending recent techniques developed for interventional inference to the realm of counterfactual inference, we offer techniques that   reduce the cost of computating bounds for probabilities of causation. These results, detailed in Section \ref{section:Approaches}, streamline the calculation and enhance the feasibility of applying these metrics in real-world settings.

Our methodological contributions center around the  use of probabilities of causation within the practical domain of Root Cause Analysis (RCA). Indeed, many practical circumstances call for a search for causes: for instance, we may search for the cause of an illness in a patient, or the cause of a service malfunction in a banking transaction. In general, an abnormal observation of a target quantity might admit several causes. For instance, a service outage may be due to a memory leak that was caused by a buggy security upgrade. 
%This memory leak explains the anomaly: had it not happened, the service would be fine. 
%One might ask whether this cause can alone be responsible for the outage, or whether an additional cause, such as energy failure, must be sought. 
The memory leak   fails to be a {\em root} cause as it is in turn  caused by the buggy upgrade. 

Traditional RCA methods often struggle under conditions of limited observability, where latent factors confound observable metrics. Probabilities of causation, with their intrinsic ability to capture counterfactual scenarios, provide a rigorous metric aligned precisely with the fundamental queries of RCA: assessing whether an observed failure would persist without a candidate cause or would necessarily occur when the cause is present. We propose a novel RCA methodology leveraging probabilities of causation not merely as isolated indicators but as core metrics for systematically comparing candidate root causes. 

The methodological proposal involves employing these probabilities to score and rank entire causal paths within a DAG, rather than isolated variables alone. This approach, detailed in Section \ref{section:Methodology}, allows identification not only of the direct root causes but also of the complete causal narrative leading to an incident. Our empirical results, conducted using the microservices domain, demonstrate the effectiveness of this causal-path scoring technique, providing clear, actionable insights into complex incident chains. 

Thus, this paper makes two complementary  contributions: it advances theoretical computation of probabilities of causation, and it introduces a practical RCA methodology that exploits these probabilities to find causal paths under realistic conditions of limited observability. After reviewing essential background material in Section \ref{section:Background}, we elaborate on computational improvements in Section \ref{section:Approaches}, then we present our methodological proposal in Section \ref{section:Methodology} and, finally, we discuss its experimental validation in Section \ref{section:Experiments}. We summarize our findings and suggest future research  in Section~\ref{section:Conclusion}.

\section{Causal Inference and Structural Causal Models}
\label{section:Background}

%In this section we review relevant elements of causal inference and root cause analysis.

%\subsection{Causal Inference with Structural Causal Models}

A \emph{Structural Causal Model} (SCM) is a tuple $(\mathcal{G},\mathbf{V},\mathbf{U},\mathcal{F},\Pr(\mathbf{U}))$ whose elements are as follows.  $\graph{G}$ is a directed graph whose node set is $\mathbf{V} \cup \mathbf{U}$, where $\mathbf{V}$ are  inner nodes called \emph{endogenous} and $\mathbf{U}$ are root nodes called \emph{exogenous}.  $\mathcal{F}$ is a set of functions $f: \val(\pa{X}) \rightarrow \val(X)$, each one of them called a \emph{mechanism}, one mechanism for each endogenous variable $X$ in $\mathcal{G}$, with $\pa{X}$ denoting the parents of $X$ in the directed graph.
Finally, $\Pr(\mathbf{U})$ is a joint probability distribution over the exogenous random variables $\mathbf{U}$. 

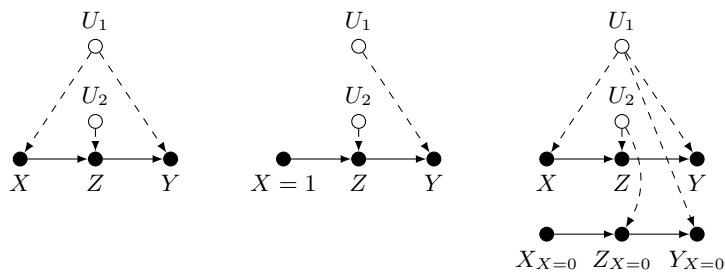
\begin{figure}[t] 
\centering
\begin{tikzpicture}
      \tikzstyle{endo} = [draw,circle,fill=black, inner sep=0.2em]
      \tikzstyle{exo} = [draw,circle,fill=white,inner sep=0.2em]
      \begin{scope}
      \node[endo,label=below:$X$] (X) at (0,0) {}; 
      \node[endo,label=below:$Z$] (Z) at (1,0) {}; 
      \node[endo,label=below:$Y$] (Y) at (2,0) {}; 
      \node[exo,label=above:$U_1$] (U1) at (1,1.5)  {}; 
      \node[exo,label=above:$U_2$] (U2) at (1,0.5)  {}; 

      \foreach \x/\y in {U1/X,U1/Y,U2/Z} {
          \draw[->,>=latex,dashed] (\x) -- (\y);
      }
      \foreach \x/\y in {X/Z,Z/Y} {
          \draw[->,>=latex] (\x) -- (\y);
      }
      \end{scope}
      \begin{scope}[xshift=3.5cm]
      \node[endo,label=below:${X=1}$] (X) at (0,0) {}; 
      \node[endo,label=below:$Z$] (Z) at (1,0) {}; 
      \node[endo,label=below:$Y$] (Y) at (2,0) {}; 
      \node[exo,label=above:$U_1$] (U1) at (1,1.5)  {}; 
      \node[exo,label=above:$U_2$] (U2) at (1,0.5)  {}; 

      \foreach \x/\y in {U1/Y,U2/Z} {
          \draw[->,>=latex,dashed] (\x) -- (\y);
      }
      \foreach \x/\y in {X/Z,Z/Y} {
          \draw[->,>=latex] (\x) -- (\y);
      }
      \end{scope}
      \begin{scope}[xshift=7cm]
      \node[endo,label=below:$X$] (X) at (0,0) {}; 
      \node[endo,label=below:$Z$] (Z) at (1,0) {}; 
      \node[endo,label=below:$Y$] (Y) at (2,0) {}; 
      \node[endo,label=below:$X_{X=0}$] (Xu) at (0,-1) {}; 
      \node[endo,label=below:$Z_{X=0}$] (Zu) at (1,-1) {}; 
      \node[endo,label=below:$Y_{X=0}$] (Yu) at (2,-1) {}; 
      \node[exo,label=above:$U_1$] (U1) at (1,1.5)  {}; 
      \node[exo,label=above:$U_2$] (U2) at (1,0.5)  {}; 

      \foreach \x/\y in {U1/X,U1/Y,U2/Z} {
          \draw[->,>=latex,dashed] (\x) -- (\y);
      }
      \foreach \x/\y in {X/Z,Z/Y,Zu/Yu,Xu/Zu} {
          \draw[->,>=latex] (\x) -- (\y);
      }
      \draw[->,>=latex,dashed] (U1) edge  (Yu); 
      \draw[->,>=latex,dashed] (U2) edge[bend left] (Zu); 
      \end{scope}  
\end{tikzpicture}
\caption{Left: a simple  SCM, capturing Pearl's Smoking/Tar/Cancer example \cite{primer}.
Middle: the intervened graph, where the edge from $U_1$ to $X$ is removed, while $X$ is fixed at $1$.
Right: a twin network where the counterfactual ``world'' is generated by $X=0$.}
\label{figure:SCM}
\end{figure}

We adopt a few assumptions about an SCM in this work: 
(i) the model is \emph{quasi-Markovian}, meaning that each endogenous variable $V_i$ has at most one exogenous parent; 
(ii) the support of the endogenous variables is finite;  
(iii) the exogenous variables are independent; and 
(iv) the graph $\graph{G}$ is acyclic. 
Assumptions (iii) and (iv) are quite common in causal inference, as they allow us to interpret the paths in the graph in terms of probabilistic (in)dependences. 
%Assumption (ii) is also not limiting, as any variable can be discretized (with some loss of information). 
Assumption (i) is arguably the most restrictive, but still covers a sufficiently large class of models of interest. Relaxing Assumption (i) makes the subsequent optimization problems much harder, and often practically intractable.

Figure \ref{figure:SCM} (left) depicts a quasi-Markovian SCM with three endogenous variables and two exogenos variables.

% Intuitively, $\mathbf{Y}_{\mathbf{X}=\mathbf{x}}$ represents the distribution of $\mathbf{Y}$ had $\mathbf{X}$ been intervened upon to be fixed at $\mathbf{x}$ (i.e., a counterfactual). This is typically not directly measurable.
% Another common notation for interventions is the ``do'' calculus one, introduced by Pearl \cite{causality}; the subscript notation just introduced is more appropriate to the kinds of calculations we consider in this paper. 

A \emph{confounded component} (for short, c-component) of a graph $\graph{G}$ is the set of endogenous nodes of a connected component of the undirected graph obtained by discarding edges whose endpoints are endogenous variables, % (i.e., $\graph{G}_{\underline{\mathbf{V}}}$), 
then dropping direction of the remaining edges \cite{Tian2002}. 
We say that an exogenous variable is associated with an endogenous variable if they appear in the same connected component of the previous undirected graph.
For a quasi-Markovian SCM, every c-component is associated with exactly one exogenous node.

In Figure \ref{figure:SCM} we have two c-components: $\{X,Y\}$ (associated with $U_1$), and $\{Z\}$ (associated with $U_2$).

For any endogenous variable $V$ in c-component $\mathbf{C}$, denote by $\mathbf{W}_V$ the endogenous variables that are topologically smaller than $V$ in the union of variables in $\mathbf{C}$ and their endogeneous parents.
Tian \cite{Tian2002} showed that the probability distribution of the endogenous random variables in quasi-Markovian SCMs factorizes as a product over c-components:
% \begin{equation}\label{eq:marginal}
% \Pr(\mathbf{V}) = \prod_{\mathbf{C}} \Pr(\mathbf{C}|\pa{\mathbf{C}}),
% \end{equation}
\begin{equation} \label{eq:marginal}
\Pr(\mathbf{V})  
% =\prod_{\mathbf{C}} %\Pr(\mathbf{C} | \Do{Pa'(\mathbf{C}))}) 
%\Pr(\mathbf{C} | \Do{\mathbf{V}\setminus \mathbf{C}})
= \prod_\mathbf{C} \prod_{V \in \mathbf{C}} \Pr(V|\mathbf{W}_{V}) .
\end{equation}
%where $\Pa'(\mathbf{C})$ denotes the endogenous parents of variables in $\mathbf{C}$ which are not in $\mathbf{C}$.
For any c-component $\mathbf{C}$ associated with exogenous $U$,
Expression~(\ref{eq:marginal}) leads to the following system of linear equalities:
\begin{equation} \label{eq:linearconstraint}
%\Pr(\mathbf{C}|\Do{\pa{\mathbf{C}}\setminus \mathbf{C}}) = 
     \prod_{V \in \mathbf{C}} \Pr(V|\mathbf{W}_V)   
     %\Pr(\mathbf{C}|\pa{\mathbf{C}}) 
     = \sum_{u: f_V(\pa{V})=V, V \in \mathbf{C}} \Pr(U = u) ,
\end{equation}
where the summation is over the values of $U$ that are consistent with the configurations of $\mathbf{C}$ and $\pa{\mathbf{C}} \cap \mathbf{V}$.

Suppose the cardinalities of exogenous variables are not known.
When endogenous variables are categorical, one can always produce categorical exogenous variables by a process called
%A Partially Specified SCM (PSCM, for short) consists of the tuple $(\mathcal{G},\mathbf{V},\mathbf{U},\mathcal{F})$; that is, it is a SCM that lacks the specification of the exogenous variables distribution. 
%When endogenous random variables are categorical, one can always extend a causal graph into a PSCM by a process know as 
\emph{canonicalization} \cite{zhang-tiam-bareinboim}, which, in essence, enumerates all possible mechanisms as latent variables.
For quasi-Markovian graphs, canonicalization reduces every exogenous variable to a categorical random variable whose state space has cardinality $\prod_{V \in \mathbf{C}}|\val(V)|^{|\val(\pa{V})|}$, where $\mathbf{C}$ is the corresponding c-component. 
Each value $u \in \val(U)$ specifies a mechanism $f_u: \pa{V} \cap \mathbf{V} \rightarrow V$ for each $V$ in the c-component.
Thus, we assume without loss of generality in the rest that every exogenous variable is categorical.

% The goal of causal inference is to estimate expressions involving such probabilities using the constraints shared by the non-intervened and intervened SCMs.

% Interventional reasoning is a relatively simple setting within causal inference. 
% More sophisticated scenarios are connected with {\em counterfactual} reasoning, summarized later when we discuss probabilities of causation. 

 A simple \emph{intervention} (a.k.a.\ treatment, exposure, action) $\Do{\mathbf{X}=\mathbf{x}}$, for $\mathbf{X} \subseteq \mathbf{V}$, modifies an SCM by substituting each mechanism $f_X$, $X \in \mathbf{X}$, with the constant value $x$ consistent with $\mathbf{x}$, thus obtaining a new SCM where the intervened variables have no parent.
This   intervened model induces a new (post-intervention) distribution over the variables, denoted by $\Pr(\mathbf{V}|\Do{\mathbf{X}=\mathbf{x}})$.
Given a   variable $Y \in \mathbf{V}$, a \emph{potential outcome} $Y_{\mathbf{X}=\mathbf{x}}(\mathbf{u})$ is the value that \emph{unit} $Y(\mathbf{u})$ would obtain had $\mathbf{X}$ been intervened upon to be $\mathbf{x}$.\footnote{Note: by construction, once the values of the exogenous random variables are fixed at some $\mathbf{U}=\mathbf{u}$, the endogenous variables become deterministic.} The probability distribution of $Y_{\mathbf{X}=\mathbf{x}}$ is then the probability induced by the distribution of the exogenous random variables. 
This is the same marginal distribution one would obtain had we generated data from the intervened SCM $\Do{\mathbf{X}=\mathbf{x}}$.
The same definitions extend to any subset of random variables $\mathbf{Y} \subseteq \mathbf{V}$, giving us sets of counterfactual random variables $\mathbf{Y}_{\mathbf{X}=\mathbf{x}}$ with joint marginal distribution $\Pr(\mathbf{Y}_{\mathbf{X}=\mathbf{x}})$.

Figure \ref{figure:SCM} (middle) depicts the intervened version of the SCM in Figure \ref{figure:SCM} (left), where $X$ is set to $1$.

A \emph{counterfactual inference} is more intricate.
Given some fixed (factual) observation $\mathbf{E}=\mathbf{e}$, the notation $\Pr(\mathbf{Y}_{\mathbf{X}=\mathbf{x}} | \mathbf{E}=\mathbf{e})$ denotes the distribution of $\mathbf{Y}_{\mathbf{X}=\mathbf{x}}$ induced by $\Pr(\mathbf{U}|\mathbf{E}=\mathbf{e})$ \cite{causality}. That is, the observation is taken before the intervention is realized.
%This reads as ``the distribution that $\mathbf{Y}$ would have had $\mathbf{X}$ been set to $\mathbf{x}$, limited to the subpopulation where $\mathbf{E}=\mathbf{e}$''.
For example, we might ask for  $\Pr(\mathbf{Y}_{\mathbf{X}=\mathbf{x}_0}=\mathbf{y}_0|\mathbf{X}=\mathbf{x}_1,\mathbf{Y}=\mathbf{y}_1)$,   the probability of   $\mathbf{Y}$ taking on some values $\mathbf{y}_0$ had we intervened upon $\mathbf{X}$ to be $\mathbf{x}_0$ when our actual observation was that $\mathbf{Y}=\mathbf{y}_1$ and $\mathbf{X}=\mathbf{x}_1$.

{\em Twin} networks are often used to compute counterfactual quantities \cite{primer}. A twin network is produced by replicating the endogenous variables and leaving the new variables with the same exogenous parents (the exogenous variables are not replicated). For instance, Figure \ref{figure:SCM} (right) shows the twin network related to Figure \ref{figure:SCM} (left) so as to compute say $\Pr(Y_{X=0}=0|Y=1)$.
%There is a replicated set of endogenous variables per distinct interventions.

\section{Computing the Probabilities of Causation}\label{section:Approaches}

Suppose we have, as usual in causal modeling, a given SCM and an {\em input distribution} $\hat{\Pr}$ over the endogenous variables $\mathbf{V}$. The latter distribution $\hat{\Pr}$ might come from a large dataset of observed endogenous variables. The assumption here is that endogenous variables are observed and exogenous variables are not. 

In looking for causes of a particular observation in this SCM, we can use three quantities based on counterfactual reasoning \cite{causality}. The first is the {\em probability of necessity} (PN) of a cause $X$ for an effect $Y$, defined as:
\[
\mathrm{PN} = \Pr(Y_{X=0}=0|X=1,Y=1).
\]
%PN is the probability that $Y=1$ 
%would fail to happen if $X=0$ were enforced,
%given that $X=1$ and $Y=1$ actually happened.
That is, PN tries to capture the degree to which $X=1$ is necessary for $Y=1$ to happen. 
The {\em probability of sufficiency} (PS) is then:
\[
\mathrm{PS} = \Pr(Y_{X=1}=1|X=0,Y=0).
\]
That is, PS tries to capture the degree to which
$X=1$ is sufficient for $Y=1$ to occur.
Finally, the {\em probability of necessity and sufficiency} (PNS) is:
\[
\mathrm{PNS} =  \Pr(Y_{X=1}=1,Y_{X=0}=0).
\]

These three quantities can be computed by replicating the variables in $\mathbf{V}$, often referred to as ``factual'' variables, so that each intervention produces a fresh set of ``counterfactual'' variables. That is, to compute the PN, we have $\mathbf{U}$, $\mathbf{V}$ and $\mathbf{V}_{X=0}$, and the latter is associated with the same mechanisms associated with $\mathbf{V}$ except for the intervened $X$. 
%Twin graphs can be used to compute PN and PS. 

In the context of root cause analysis, Oliveira et al.\ \cite{Oliveira2022} proposed interventional quantities as proxies for the previous definitions. 
Inspired by their proposals, we   define, for a tentative cause $X$ and an effect of interest $Y$:
\begin{itemize}
\item The {\em weak probability of necessity} (w-PN) is $\Pr(Y=0|\mathrm{do}(X=0))$.
\item The {\em weak probability of sufficiency} (w-PS) is $\Pr(Y=1|\mathrm{do}(X=1))$.
%\item The {\em weak probability of necessity and sufficiency} (w-PNS) is \alpha \mbox{w-PN} + (1-\alpha) \mbox{w-PS}$ for some $\alpha\in[0,1]$.
\end{itemize}

Usually, the cardinality of exogenous variables is not known; as noted in Section \ref{section:Background}, the canonicalization procedure can then be employed to build the exogenous variables. 
%The resulting cardinalities of exogenous variables have been presented in Section \ref{section:Background}.

There are interventional probabilities that can be   identified using previous results (that is, a single real number is obtained from graph and  input distribution). For instance, for the graph in Figure \ref{figure:SCM} (left), the w-PS can be precisely computed using the front-door criterion \cite{primer}. However, in general, interventional and counterfactual probabilities cannot be precisely identified, and we must resort to computing tight lower and upper bounds for them. Hence we must focus, for any   $Q$, either on its {\em lower value} $\underline{Q}=\inf Q$, or its {\em upper value} $\overline{Q} = \sup Q$. 

Previous work has shown that the computation of lower and upper values for interventional quantities, given a canonicalized SCM an an input distribution, leads to minimization/maximization of a multilinear objective function subject to linear constraints \cite{zaffalon2024}. For a quasi-Markovian SCM,  the degree of the multilinear objective function is, in principle, equal to the number of exogenous variables. 
Hence the feasible region is closed and the upper probabilities are obtained by maximization while the lower probabilities are obtained by minimization \cite{zhang-tiam-bareinboim}.  

In a recent breakthrough,
Shridharam and Iyengar \cite{shridharan23icml} have shown that interventional quantifies require a multilinear objective function whose degree is {\em only} equal to the number of distinct intervened c-components. Hence, tight bounds for interventions in a single c-component require only linear programs!
Such results exploit properties of the input distribution to remove exogenous variables from the calculations.

Alas, these previous results by Shridharan and Iyengar do {\em not} apply to counterfactual quantities such as PN or PS. 
This can be already seen in connection with the simple network in Figure \ref{figure:SCM} (left). Suppose we wish to compute $\underline{\mbox{PN}}$; to do so, we resort to the twin network in Figure \ref{figure:SCM} (right). However, in that network we cannot remove $U_2$; there is no corresponding simplification to be made to the ensuing multilinear objective function. 

Is it possible to remove exogenous variables while computing probabilities of causation PN, PS, PNS? We now present novel results regarding this question, in essence by combining and extending previous efforts. 

We start by noting that the PN and the PS can be written as
\begin{equation}
\label{equation:FractionalPNPS}
\mathrm{PN} = \frac{\Pr(Y_{X=0}=0,X=1,Y=1)}{\hat{\Pr}(X=1,Y=1)}, \;\;
\mathrm{PS} = \frac{\Pr(Y_{X=1}=1,X=0,Y=0)}{\hat{\Pr}(X=0,Y=0)},
\end{equation}
where numerators are computed using the twin network, and where
denominators depend only on the factual variables whose distribution is the input $\hat{\Pr}$.
The latter fact let us avoid fractional objective functions while computing PN or PS.

The next move is to use the results of Shpitser and Pearl \cite{Shpitser2007} that allow us to simplify the twin networks. In fact, Shpitser and Pearl studied general settings in which many different counterfactual ``worlds'' must be considered, moving away from the two-world setting of twin networks. Their {\em counterfactual graphs} have as many copies of the endogenous variables as there are different  interventions. More importantly, a counterfactual graph removes variables that are not necessary, in some cases by merging variables that are actually identical.
The next example illustrates the properties of counterfactual graphs in an interventional setting. 

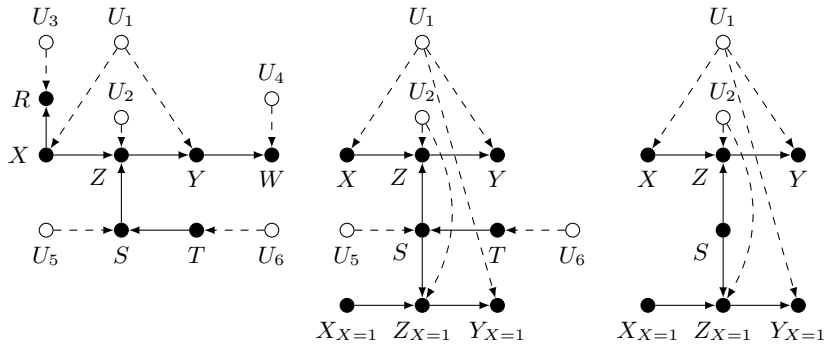
\begin{figure}[t] 
\centering
\begin{tikzpicture}
      \tikzstyle{endo} = [draw,circle,fill=black, inner sep=0.2em]
      \tikzstyle{exo} = [draw,circle,fill=white,inner sep=0.2em]
      \begin{scope}
      \node[endo,label=left:$X$] (X) at (0,0) {}; 
      \node[endo,label=below left:$Z$] (Z) at (1,0) {}; 
      \node[endo,label=below:$Y$] (Y) at (2,0) {}; 
      \node[endo,label=below:$S$] (S) at (1,-1) {}; 
      \node[endo,label=below:$T$] (T) at (2,-1) {}; 
      \node[endo,label=left:$R$] (R) at (0,0.75) {}; 
      \node[endo,label=below:$W$] (W) at (3,0) {}; 
    
      \node[exo,label=above:$U_1$] (U1) at (1,1.5)  {}; 
      \node[exo,label=above:$U_2$] (U2) at (1,0.5)  {};
      \node[exo,label=above:$U_3$] (U3) at (0,1.5)  {}; 
      \node[exo,label=above:$U_4$] (U4) at (3,0.75)  {}; 
      \node[exo,label=below:$U_5$] (U5) at (0,-1)  {}; 
      \node[exo,label=below:$U_6$] (U6) at (3,-1)  {}; 
      
      \foreach \x/\y in {U1/X,U1/Y,U2/Z,U3/R,U4/W,U5/S,U6/T} {
          \draw[->,>=latex,dashed] (\x) -- (\y);
      }
      \foreach \x/\y in {X/Z,Z/Y,Y/W,X/R,S/Z,T/S} {
          \draw[->,>=latex] (\x) -- (\y);
      }
      \end{scope}
      \begin{scope}[xshift=4cm]
      \node[endo,label=below:$X$] (X) at (0,0) {}; 
      \node[endo,label=below left:$Z$] (Z) at (1,0) {}; 
      \node[endo,label=below:$Y$] (Y) at (2,0) {}; 
      \node[endo,label=below left:$S$] (S) at (1,-1) {}; 
      \node[endo,label=below:$T$] (T) at (2,-1) {}; 

      \node[endo,label=below:$X_{X=1}$] (X1) at (0,-2) {}; 
      \node[endo,label=below:$Z_{X=1}$] (Z1) at (1,-2) {}; 
      \node[endo,label=below:$Y_{X=1}$] (Y1) at (2,-2) {};
      
      \node[exo,label=above:$U_1$] (U1) at (1,1.5)  {}; 
      \node[exo,label=above:$U_2$] (U2) at (1,0.5)  {};
      \node[exo,label=below:$U_5$] (U5) at (0,-1)  {}; 
      \node[exo,label=below:$U_6$] (U6) at (3,-1)  {}; 
      
      \foreach \x/\y in {U1/X,U1/Y,U2/Z,U5/S,U6/T} {
          \draw[->,>=latex,dashed] (\x) -- (\y);
      }
      \foreach \x/\y in {X/Z,Z/Y,S/Z,T/S} {
          \draw[->,>=latex] (\x) -- (\y);
      }
      \foreach \x/\y in {U1/Y1} {
          \draw[->,>=latex,dashed] (\x) -- (\y);
      }
      \draw[->,>=latex,dashed] (U2) edge[out=-60,in=60] (Z1);
      \foreach \x/\y in {X1/Z1,Z1/Y1,S/Z1} {
          \draw[->,>=latex] (\x) -- (\y);
      }
      \end{scope}
      \begin{scope}[xshift=8cm]
      \node[endo,label=below:$X$] (X) at (0,0) {}; 
      \node[endo,label=below left:$Z$] (Z) at (1,0) {}; 
      \node[endo,label=below:$Y$] (Y) at (2,0) {}; 
      \node[endo,label=below left:$S$] (S) at (1,-1) {};  

      \node[endo,label=below:$X_{X=1}$] (X1) at (0,-2) {}; 
      \node[endo,label=below:$Z_{X=1}$] (Z1) at (1,-2) {}; 
      \node[endo,label=below:$Y_{X=1}$] (Y1) at (2,-2) {};
      
      \node[exo,label=above:$U_1$] (U1) at (1,1.5)  {}; 
      \node[exo,label=above:$U_2$] (U2) at (1,0.5)  {}; 
      
      \foreach \x/\y in {U1/X,U1/Y,U2/Z} {
          \draw[->,>=latex,dashed] (\x) -- (\y);
      }
      \foreach \x/\y in {X/Z,Z/Y,S/Z} {
          \draw[->,>=latex] (\x) -- (\y);
      }
      \foreach \x/\y in {U1/Y1} {
          \draw[->,>=latex,dashed] (\x) -- (\y);
      }
      \draw[->,>=latex,dashed] (U2) edge[out=-60,in=60] (Z1);
      \foreach \x/\y in {X1/Z1,Z1/Y1,S/Z1} {
          \draw[->,>=latex] (\x) -- (\y);
      }
      \end{scope}
\end{tikzpicture}
\caption{Left: a quasi-Markovian SCM.
Middle: the counterfactual graph for cause $X$ and effect $Y$.
Right: the reduced counterfactual graph (applying Theorem \ref{theorem:Reduce}).
}
\label{figure:LargerSCM}
\end{figure}

\begin{example}\label{example:LargerSCM}
Consider the quasi-Markovian SCM in Figure \ref{figure:LargerSCM} (left). 
A twin network might be built and, based on it, a multilinear objective function for minimization of PN for cause $X$ and effect $Y$ would have degree 6.
The counterfactual graph is shown in Figure \ref{figure:LargerSCM} (middle), taking intervention $X=1$. 
Note that $R$ and $W$ (and associated exogenous variables) are removed as they cannot affect the computation of interest (they are not ascendants of relevant variables).
Also, the duplicates of $S$ and $T$ are not included in the counterfactual graph, because they {\em must} be identical to factual $S$ and $T$ as they are associated with identical mechanisms and no intervened variables. 
$\Box$
\end{example}

A counterfactual graph can lead to a substantial simplification: in this example, we moved from a multilinear program with degree 6 to one with degree 4. Can we remove additional exogenous variables?
Under some conditions, yes:

\begin{theorem}\label{theorem:Reduce}
Suppose we have a set of endogenous variables $\mathbf{Z}$ in a counterfactual graph such 
that no variable in  $\mathbf{Z}$  is a descendant 
of the counterfactual target $Y_{X=x}$ (where $x$ can be zero or one, depending on whether PN or PS is to be computed), the factual intervened $X$ or the factual target $Y$; moreover, suppose 
$\mathbf{Z}$ d-separates, given the factual intervened $X$ and the factual target $Y$, all its ascendants from $Y_{X=x}$.
Then the computation of lower and upper versions of PN and of PS
are not affected if all ascendants of $\mathbf{Z}$ (including exogenous variables) are
removed from the counterfactual graph, and the marginal distribution of $\mathbf{Z}$ is
then be set at  $\hat{\Pr}(\mathbf{Z})$ (from the input distribution $\hat{\Pr}$).
\end{theorem}

\begin{proof}
Let $\mathbf{W}$ denote the ascendants of $\mathbf{Z}$ not in $\mathbf{Z}$.
% \[
% \sum_{\mathbf{Z},\mathbf{W}} \Pr(Y_{X=x}=y|X=1-x,Y=1-x,\mathbf{Z},\mathbf{W}) \Pr(\mathbf{Z},\mathbf{W}|X=1-x,Y=1-x) .
% \]
Since $\mathbf{Z}$ d-separates $Y_{X=x}$ and $\mathbf{W}$ given $X$ and $Y$, we can compute $\Pr(Y_{X=x}=y|X=1-x,Y=1-x)$ as
\[
\sum_{\mathbf{Z}} \Pr(Y_{X=x}=y|X=1-x,Y=1-x,\mathbf{Z}) \sum_{\mathbf{W}} \Pr(\mathbf{Z},\mathbf{W}|X=1-x,Y=1-x) .
\]
Now the conditional probability distribution over $\mathbf{Z},\mathbf{W}$ is invariant to the intervention (as it refers only to pre-treatment variables).
Hence, it can be computed from the input  distribution (i.e., it is identifiable) $\Pr(\mathbf{Z}|X=1-x,Y=1-x)$.
This is the same estimand one obtains if we start from a modified model where $\mathbf{W}$ is removed and $\mathbf{Z}$ is replaced by a single node whose marginal is $\hat{\Pr}(\mathbf{Z})$.
% If $\mathbf{Z}$ satisfies the stated assumptions, we can apply a variable elimination
% algorithm to the ascendants of $\mathbf{Z}$, removing them all but perhaps introducing
% dependencies among variables in $\mathbf{Z}$. After we remove all ascendants of
% $\mathbf{Z}$, we can proceed writing down the expression of
% $\Pr(Y_{X=x}=y|X=1-x,Y=1-x)$, for appropriate
% $x$ and $y$, by running variable elimination symbolically. 
% In doing so, we can use $\Pr(\mathbf{Z})$ whenever the variables in $\mathbf{Z}$ are to be eliminated. 
% But $\Pr(\mathbf{Z})$ is actually given by the input distribution as
% $\hat{\Pr}(\mathbf{Z})$, so we can use the latter distribution when writing down the
% objective function without actually manipulating the exogenous variables that are
% ascendants of $\mathbf{Z}$.
$\Box$
\end{proof}

This theorem resembles simplifications by Shridharan and Iyengar \cite{shridharan23icml}; however, they do not deal with counterfactual variables. And their results can be adapted to prove:

\begin{corollary}\label{collorary:Reduce}
Suppose a counterfactual graph is reduced using Theorem \ref{theorem:Reduce}.
Tight bounds for PN or PS can be computed by minimizing/maximizing a multilinear objective function whose degree is equal to the number of exogenous variables in the reduced graph, subject to linear constraints. 
\end{corollary}

\begin{example}
Consider again Example \ref{example:LargerSCM}.
Using Theorem \ref{theorem:Reduce}, we can reduce the counterfactual graph in Figure \ref{figure:LargerSCM} (middle) to the graph in Figure \ref{figure:LargerSCM} (right);
  $U_5$ and $U_6$ have been removed. Bounds on PN or on PS can be obtained minimizing/maximizing appropriate multilinear objective functions of degree 2 (instead of degree 4 using the counterfactual graph), as shown in Example \ref{example:CompleteExample}.
$\Box$
\end{example}

Now consider the computation of PNS. This is somewhat different from PN or PS, as now we have two interventions (thus producing two counterfactual ``worlds''). However, the two counterfactual  variables in 
$\Pr(Y_{X=1}=1,Y_{X=0}=0)$ are not descendants of the factual endogenous variables, hence the latter variables are not present in the counterfactual graph. 
Moreover, the   simplifications to the counterfactual graph that are
described by Theorem~\ref{theorem:Reduce} can be adapted to the
computation of PNS as follows:

\begin{theorem}
Suppose we have a set of endogenous variables $\mathbf{Z}$ in a counterfactual graph such 
that no variable in  $\mathbf{Z}$  is a descendant 
of counterfactual variables $Y_{X=1}$ and $Y_{X=0}$; moreover, suppose 
 $\mathbf{Z}$ d-separates all its ascendants from $Y_{X=1}$ and $Y_{X=0}$.
Then the computation of lower and upper versions of PNS
are not affected if all ascendants of $\mathbf{Z}$ (including exogenous variables) are
removed from the counterfactual graph, and the marginal distribution of $\mathbf{Z}$ is
then be set at  $\hat{\Pr}(\mathbf{Z})$ (from the input distribution $\hat{\Pr}$).
\end{theorem}

%The proof of this theorem follows from the same steps employed in the proof of %Theorem \ref{theorem:Reduce}, replacing the variables $Y_{X=x}$, $X$ and $Y$ %employed there by $Y_{X=1}$ and $Y_{X=0}$. 

A parallel to Corollary \ref{collorary:Reduce} can then be derived (that is, tight bounds are obtained minimizing/maximizing a multilinear expression whose degree is the number of exogenous variables in a reduced graph). 

\begin{example}\label{example:CompleteExample}
Consider again  Example \ref{example:LargerSCM} and suppose we wish to compute $\underline{\mbox{PN}}$. As noted previously, we have to process the graph in Figure \ref{figure:LargerSCM} (right) to obtain a bilinear objective function (that is, a multilinear expression of degree 2) subject to linear constraints.
Consider first the linear constraints, produced out of Expression (\ref{eq:linearconstraint}). 
Writing these linear constraints down requires specifying an encoding for the mechanisms (as done by Balke and Pearl in their original discussion of interventional bounds \cite{balke-pearl}). 
Start with the c-component associated with $U_1$ to impose constraints on $\Pr(U_1)$. Note that $U_1$ has 8 values; 
denote $\Pr(U_1=j)$ by $q_{1,j}$ with $j \in \{0,\dots,7\}$. 
Suppose we take $X=0$ when $j\in\{0,1,2,3\}$ and $X=1$ otherwise; and we take
$Y=0$ when $j=0$ or $j=4$, $Y=Z$ when $j=1$ or $j=5$, $Y=1-Z$ when $j=2$ or $j=6$, and $Y=1$ when $j=3$ or $j=7$. 
We then have 8 constraints, one per configuration of $X,Y,Z$, using the template
\[
{\textstyle \sum_j} q_{1,j} [\![ f_X(j) = x]\!] [\![ f_Y(z,j)=y]\!] = g_1(xyz),
%\hat{\Pr}(Y=y|X=x,Z=z)\hat{\Pr}(X=x),
\]
where $[\![\theta]\!]$ is the indicator function of equality $\theta$ (that is, $1$ is $\theta$ holds, $0$ otherwise), and
$g_1(xyz) := \hat{\Pr}(Y=y|X=x,Z=z)\hat{\Pr}(X=x)$.
% The constraints are:
% \[
% \begin{array}{c}
% q_{1,0}+q_{1,1}=g_1(000), \qquad
% q_{1,0}+q_{1,2}=g_1(001), \\
% q_{1,2}+q_{1,3}=g_1(010), \qquad
% q_{1,1}+q_{1,3}=g_1(011), \\
% %%
% q_{1,4}+q_{1,5}=g_1(100), \qquad
% q_{1,4}+q_{1,6}=g_1(101), \\
% q_{1,6}+q_{1,7}=g_1(110), \qquad
% q_{1,5}+q_{1,7}=g_1(111).
% \end{array}
% \]
Now consider the c-component associated with $U_2$ to impose constraints on $\Pr(U_2)$. Note that $U_2$ has 16 values, and adopt $q_{2,k} := \Pr(U_2=k)$.  
Write $k$ as a binary number with bits $b_0b_1b_2b_3$, and take the mechanism $f_Z(s,x,k)$ as follows: if $sx=00$, then the mechanism $f_Z$ yields the first bit $b_0$ of $k$; if $sx=01$, then $f_Z$ yields $b_1$; if $sx=10$, then $f_Z$ yields $b_2$; and if $sx=11$, then $f_Z$ yields $b_3$. 
We then have 8 constraints, one per configuration of $S,X,Z$, using the template
\[
{\textstyle\sum_k} q_{2,k} [\![f_Z(s,x,k)=z]\!] = g_2(sxz),
\]
where $g_2(sxz) := \hat{\Pr}(Z=z|S=s,X=x)$.
% The constraints are:
% \[
% \begin{array}{r}
% q_{2,0}+q_{2,1}+q_{2,2}+q_{2,3}+q_{2,4}+q_{2,5}+q_{2,6}+q_{2,7}=g_2(000), \\
% q_{2,8}+q_{2,9}+q_{2,10}+q_{2,11}+q_{2,12}+q_{2,13}+q_{2,14}+q_{2,15}=g_2(001), \\
% q_{2,0}+q_{2,1}+q_{2,2}+q_{2,3}+q_{2,8}+q_{2,9}+q_{2,10}+q_{2,11}=g_2(010), \\
% q_{2,4}+q_{2,5}+q_{2,6}+q_{2,7}+q_{2,12}+q_{2,13}+q_{2,14}+q_{2,15}=g_2(011), \\
% %%
% q_{2,0}+q_{2,1}+q_{2,4}+q_{2,5}+q_{2,8}+q_{2,9}+q_{2,12}+q_{2,13}=g_2(100), \\
% q_{2,2}+q_{2,3}+q_{2,6}+q_{2,7}+q_{2,10}+q_{2,11}+q_{2,14}+q_{2,15}=g_2(101), \\
% q_{2,0}+q_{2,2}+q_{2,4}+q_{2,6}+q_{2,8}+q_{2,10}+q_{2,12}+q_{2,14}=g_2(110), \\
% q_{2,1}+q_{2,3}+q_{2,5}+q_{2,7}+q_{2,9}+q_{2,11}+q_{2,13}+q_{2,15}=g_2(111).
% \end{array}
% \]
Then the value of PN, to be minimized, is the  bilinear expression:
$q_{1,1} h_1 + q_{1,2} h_2$, where
\[
\begin{array}{c}
h_1 := 
(1-\gamma)
(q_{2,4}+q_{2,5}+q_{2,6}+q_{2,7}) +
\gamma
(q_{2,1}+q_{2,5}+q_{2,9}+q_{2,13}),
\\ 
h_2 := 
(1-\gamma)
(q_{2,8}+q_{2,9}+q_{2,10}+q_{2,11})
+
\gamma
(q_{2,2}+q_{2,6}+q_{2,10}+q_{2,14}),
\end{array}
\]
with $\gamma=\Pr(S=1)$. 
A similar expression is obtained for PNS. Even though the PNS requires two interventions, it does not require conditioning --- for this reason, the reduced counterfactual graph to be processed is almost identical to the graph in Figure \ref{figure:LargerSCM} (right): the only difference is that we have $X_{X=0}$ instead of $X$, $Z_{X=0}$ instead of $Z$, and $Y_{X=0}$ instead of $Y$. 
The value of PNS, to be minimized, is the following expression (again subject to the same linear constraints described before): $(q_{1,1}+q_{1,5})h_1 + (q_{1,2}+q_{1,6})h_2$.

To illustrate, take the following randomly generated vector, containing values of
the input distribution $\hat{\Pr}(X=x,Y=y,Z=z,S=s)$ ordered as if $xyzs$ were a binary number:
$[0.30375, 0.0075, 0.03375, 0.3, 0.12, 0.0225, 0.03, 0.2025, 
  0.0675,$ $0.01, 0.0075, 0.4, 0.01, 0.01125, 0.0025, 0.10125]$.
We can use the Gurobi optimization
solver,\footnote{Gurobi: https://gurobi.com/} to easily obtain the PNS interval 
$[0.35, 0.49]$, in less than 10ms, and the PN interval $[0.175, 0.245]$, 
also in less than 10ms.
%
% \begin{tabular}{|c||c|c|c|c|c|c|c|c|} \hline
% $s \; \downarrow \; \backslash \; xyz \rightarrow$ &
% 000 & 001 & 010 & 011 &
% 100 & 101 & 110 & 111 \\ \hline \hline
% 0 & 0.30375 & 0.0075 &  0.03375 & 0.3 & 0.12 & 0.0225  & 0.03   & 0.2025 \\ \hline
% 1 & 0.0675  & 0.01   &  0.0075  & 0.4 & 0.01 & 0.01125 & 0.0025 & 0.10125\\ \hline
% \end{tabular}
$\Box$
\end{example}

% ANOTHER SIMPLIFICATION, USING EXAMPLE 1:
% The three constraints over $\Pr(U_2)$ lead to a
% credal set that is one-dimensional, a line segment in 4-dimensional space parameterized by 
% \[
% \begin{array}{c}
% t \in  [\max(0,\hat{\Pr}(Z=1|X=1)+\hat{\Pr}(Z=1|X=0)-1), \\
% \min(\hat{\Pr}(Z=1|X=1),\hat{\Pr}(Z=1|X=0))]
% \end{array}
% \]
% as follows:
% \[
% \left[ 
% \begin{array}{c}
% 1 + t - \hat{\Pr}(Z=1|X=1) - \hat{\Pr}(Z=1|X=0)  \\
% \hat{\Pr}(Z=1|X=0) - t \\
% \hat{\Pr}(Z=1|X=1) - t \\
% t
% \end{array}
% \right].
% \]

\section{Probabilities of Causation in Root Cause Analysis}

In the previous section we introduced new techniques to simplify the computation of probabilities of causation. 
We now investigate how to use those probabilities in a practical challenge, namely, in the search for root causes.

\subsection{Root Cause Analysis: A Very Brief Overview} 

An anomaly occurs when a system exhibits a set of symptoms signaling a potential failure, triggering the need to identify its underlying causes to rectify or mitigate the issue.
Finding root causes is, however, a complex task. If a module fails, and its behavior depends on a sequence of previous actions, it may difficult to find the specific root source of the failure.
%This task, however, is inherently complex. Consider a simple causal diagram with modules connected as: Module 1 $\rightarrow$ Module 2 $\rightarrow$ Module 3. If Module 3 presents anomalous behavior, Module 2 naturally emerges as a candidate cause. However, further inspection is necessary to determine whether the root issue resides in Module 2 itself or upstream in Module 1. Thus, effective root cause analysis (RCA) must not only identify anomalies but also assess the relative importance of candidate causes.

Root Cause Analysis (RCA) range from manual, diagram-based inspections to advanced automated systems leveraging statistical or causal inference techniques \cite{Wang2024}. These approaches are broadly employed in domains like epidemiology, medical diagnostics and, increasingly, system observability, particularly for microservices and distributed infrastructures \cite{Ikram2022,Mokhtari2023}. The complexity and opacity of modern systems, due to hidden components or limited monitoring, pose significant challenges for traditional RCA methods that rely on correlations between monitored signals \cite{Budhathoki2022,Wang2024}. Unobserved factors (latent confounders) frequently obscure the true causal relationships, complicating root cause identification.

Existing causal  RCA methods typically involve constructing causal graphs either from domain expertise or automated causal discovery algorithms, subsequently assigning scores to each node reflecting their contribution to the observed anomaly \cite{Ikram2022,Li2022}. Several scoring metrics have been proposed, including Shapley-style attributions derived from soft interventions, as implemented in libraries such as DoWhy. These methods yield a ranked list of candidate root causes based on their causal relevance to the incident \cite{Budhathoki2022}.
Even though there are approaches that employ causal tools, the literature is very sparse about using probabilities of causation for RCA, particularly under scenarios of limited information.

\subsection{A Proposal: Causal Narratives via Probabilities of Causation}\label{section:Methodology}

Probabilities of causation are effective in establishing to what degree a particular variable can be deemed a cause of an event. 
Yet, assessing whether a particular variable is a root cause of an incident requires considering distinctions between degrees of causation between variables.
We now propose a technique that uses probability of causation to identify what we call \emph{causal narratives}, roughly understood as a causal chain of events that connect a single root cause to an incident observed in sample data containing examples of both normal and abnormal behavior.
The root cause is thus the head of such causal chain.

% Each narrative fixes a single, \emph{ground-truth} root cause and re-estimates the
% tight analytical bounds of the five probabilities of causation-based metrics (PN, PS, PNS, weak-PN and weak-PS) for all observable variables under that assumption. 
% This produces a family of score tables, one per narrative, that encode how strongly each node would have to contribute in order for the chosen root cause to be consistent with the data generating structural causal model. 

Algorithm~\ref{alg:dfs-short} operates on probability scores, using a pruned-DFS (Depth-First Search) to rank the causal chains and assessing whether they can recover the injected root-cause narratives from the resulting probability values. A few details about the algorithm:

\vspace{.5em}
\noindent\textbf{Search Metric.}
Let \(G=(V,E)\) be a directed acyclic graph (DAG) whose edges encode
causal relations between observed variables and let
\(Y\in V\) be a distinguished \emph{incident node}.
For every \(X\in V\setminus\{Y\}\) we compute a local
\emph{probability-of-causation}-based metric $\text{PC}(X,Y)$ of choice. For instance, $\text{PC}(X,Y)$ may be chosen to refer to the Probability of Necessity (PN). %
% —
% defined by
% \(
% \text{PN}(X,Y)=
% P\!\bigl(Y_{x=1}=1,\;Y_{x=0}=0\bigr)
% \in[0,1].
% \)
% DENIS: acho que esse comentário fica estranho aqui. Talvez faça mais sentido colocar ele para justificar porque esperamos que PN tenha se saído melhor.
%Because PN admits an intuitive interpretation as
% ``causal risk contribution'', we treat it as an \emph{additive weight}
% attached to each node.

\vspace{.5em}
\noindent\textbf{Search Strategy.}
We perform a depth-first traversal that walks \emph{upstream} from the
incident node.
Whenever the absolute change
\(\Delta_{k+1}=|\,\text{PC}_{k+1}-\text{PC}_{k}\,|\)
between two consecutive ancestors
exceeds a fixed multiple \(\alpha>1\) of the current median
\(\mathcal M_k=\mathrm{median}\{\Delta_1,\dots,\Delta_{k}\}\),
we \emph{prune} the branch and declare the present node a plausible
root cause.
This rule removes spurious upstream nodes whose causal strength drops
abruptly while preserving coherent high-PC chains.
%(cf.\ Algorithm~\ref{alg:dfs-short}).

\vspace{.5em}
\noindent\textbf{Path Significance.}
For every surviving path
\(\pi=\langle X_r,\dots,X_1,Y\rangle\)
we define a linear
\emph{path-significance score} define as
$S(\pi)=w\cdot\text{PC}(X_r,Y)\;+\;
\sum_{j=1}^{r-1}\text{PC}(X_j,Y)$,
where \(w \geq 1\) is an adjustable weight that emphasizes the putative  
root cause \(X_r\).
Paths are ranked by \(S(\pi)\); the top entry yields both the
most likely root cause and its dominant causal chain.

\begin{algorithm}[t]
\caption{Root-Cause Path Ranking}
\label{alg:dfs-short}
\begin{multicols}{2}
\begin{algorithmic}[1]
\Require DAG $G$, PC scores $\text{PC}[\cdot]$, target $Y$,
        drop factor $\alpha$, root weight $w$
\Function{Rank}{$G,\text{PC},Y$}
  \State $\texttt{out}\gets[\,]$

  \Procedure{DFS}{$v,\,P,\,\Delta$}
    \If{$parents(v)=\emptyset$}\State \Call{rec}{$P$}\; \Return\EndIf
    \For{$p\in parents(v)$}
      \State $\delta\gets|\text{PC}[p]-\text{PC}[v]|$     
        \State $\beta \gets \alpha\cdot\mathrm{median}(\Delta)$ \If{$|\Delta|\ge1\land\delta>\beta$}
          \State \Call{rec}{$P\cup\{p\}$};\; 
          $\mathbf{continue}$ % \Continue 
        \EndIf
      \State \Call{DFS}{$p,\,P\cup\{p\},\,\Delta\cup\{\delta\}$}
    \EndFor
  \EndProcedure
  \Function{rec}{$P$}
    \State $s\gets w\cdot\text{PC}[P_0]+\displaystyle\sum_{x\in P\setminus\{Y\}}\text{PC}[x]$
    \State push $(P^{\text{rev}},s)$ into \texttt{out}
  \EndFunction
  \State \Call{DFS}{$Y,[Y],[\,]$}
  \State \Return \texttt{out} sorted by $s$ $\downarrow$
\EndFunction
\end{algorithmic}
\end{multicols}
\end{algorithm}

% To perform inference and estimate each of the five considered metrics, across all model-narrative pairs, we employed the EMCC procedure \cite{zaffalon2024} implemented in the Python framework \texttt{BCAUSE}.\footnote{Available at 
% \url{https://github.com/PGM-Lab/bcause}.} The procedure uses a parameter learning approach to approximate the probability bounds of different probability of causation scores.

\subsection{Experiments and Evaluation: RCA for Microservices}\label{section:Experiments}

\begin{figure}[t]
\centering
\begin{tikzpicture}
  \tikzstyle{endo} = [draw,circle,fill=black, inner sep=0.2em]
  \tikzstyle{exo}  = [draw,circle,fill=white,inner sep=0.2em]

  \node[endo,label=below:{NewDeploy}]      (ND) at (0,0)  {};
  \node[endo,label=below:{MemoryLeak}]     (ML) at (2.5,0)  {};
  \node[endo,label=below:{MemUsageHigh}]   (MU) at (5,0)  {};
  \node[endo,label=below:{ServiceCrash}]   (SC) at (7.5,0)  {};
  \node[endo,label=below:{OutageIncident}] (OI) at (10,0)  {};

  \node[exo,label=left:{HeavyTraffic}]    (HT) at (6.25,1) {};

  % ----------- ARESTAS (confundidor – tracejado)-----------
  \foreach \x/\y in {HT/MU, HT/SC}{
      \draw[->,>=latex,dashed] (\x) -- (\y);
  }
  % ----------- ARESTAS (cadeia principal – sólidas) -------
  \foreach \x/\y in {ND/ML, ML/MU, MU/SC, SC/OI}{
      \draw[->,>=latex] (\x) -- (\y);
  }
\end{tikzpicture}
\caption{Model 1: Small-scale observability causal model. Dashed arrows indicate latent confounding by a surge in traffic.}
\label{figure:incident-dag}
\end{figure}
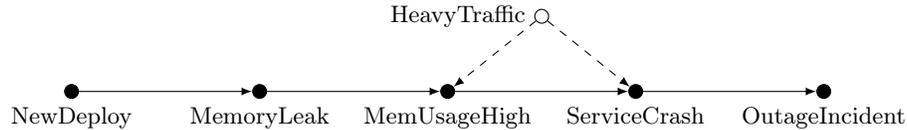

We describe experiments that assess the performance of our proposed metrics in the context of RCA for 
%
%setup is to assess the effectiveness of using probabilities of causation as metrics for root cause analysis (RCA), specifically within 
microservices characterized by limited information. 
Microservice outages typically stem from cascading failures, where an initial fault propagates through dependent components. Effective RCA must therefore accurately trace the causal pathways that lead to such incidents.

We investigate metrics based on PN, PS, PNS, w-PN and w-PS as defined in Section \ref{section:Approaches}.
The estimation of probabilities of causation typically yields interval estimates rather than precise point values. To practically use these interval estimates, we apply heuristic strategies to reduce each interval to a single scalar value, such as Minimum (to consider the lower bound as the score, thus prioritizing the estimates of the metrics by how far from zero, of no importance, they are), Maximum (to use the upper bound of the interval as the score, thus prioritizing the estimates of the metrics by their maximum possible importance to the observed effect), Mean and Midpoint.

We designed experiments based on synthetic causal models and datasets to evaluate how effectively these metrics identify   root causes under scenarios involving latent confounders and partial observability. Each synthetic model represents a simplified but realistic microservice system, where nodes correspond to binary representations of service-level metrics (such as high latency or error occurrences, determined by specified thresholds). Given space restrictions, we focused on models that can illustrate the mechanics of our techniques, and not on models engineering to display computational gains produced by the techniques in the previous sections. 

% \begin{itemize}
% \item \textbf{PN, PS and PNS}: These metrics as they are formally defined by queries that measure strong counterfactual relationships. Specifically, PN assesses whether the absence of a cause would prevent an observed effect; PS assesses if the presence of the cause ensures the effect; and PNS evaluates a joint criterion combining necessity and sufficiency.

% \item \textbf{Weak-PN}: Defined as $P(Y=0 \mid do(X=0))$, this interventional approximation measures how effectively an intervention removing the cause prevents the incident.

% \item \textbf{Weak-PS}: Defined as $P(Y=1 \mid do(X=1))$, this measure assesses how likely an intervention introducing the cause leads directly to the incident.
% \end{itemize}

To evaluate our proposed metrics rigorously, we designed three synthetic causal models of varying complexity to simulate typical fault scenarios encountered in microservices architectures, as illustrated in Figures 3-5. Each model's   mechanisms and corresponding synthetic datasets explicitly encode known causal narratives, allowing us to objectively assess how well each causation-based metric recovers these true underlying causal relationships. 
%Through these systematically controlled experiments, we evaluated probabilities of causation as scoring methods for RCA, looking at distinct operational scenarios and objectives in real-world microservice monitoring and management.
% The corresponding data-generation procedures and scripts are provided in the Appendix.

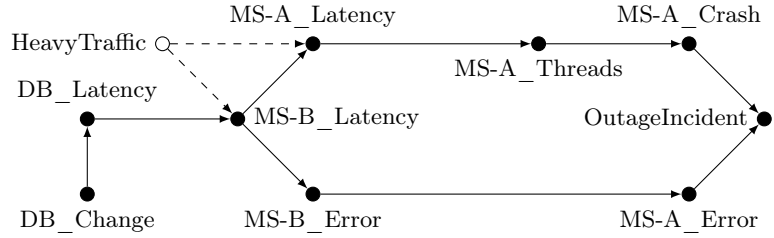
\begin{figure}[t]
\centering
\begin{tikzpicture}
  \tikzstyle{endo} = [draw,circle,fill=black, inner sep=0.2em]
  \tikzstyle{exo}  = [draw,circle,fill=white,inner sep=0.2em]

  % ---------------------- NÓS ----------------------------
  \node[endo,label=below:{DB\_Change}]           (DBC) at (2,0.5)   {};
  \node[endo,label=above:{DB\_Latency}]            (DBL) at (2,1.5)   {};
  
  \node[endo,label=right:{MS-B\_Latency}]      (SBL) at (4,1.5) {};
  
  \node[endo,label=above:{MS-A\_Latency}]      (SAL) at (5,2.5) {};
  \node[endo,label=below:{MS-A\_Threads}]  (SAT) at (8,2.5) {};
  \node[endo,label=above:{MS-A\_Crash}]            (SAC) at (10,2.5){};

  \node[endo,label=below:{MS-B\_Error}]        (SBE) at (5,0.5){};
  \node[endo,label=below:{MS-A\_Error}]        (SAE) at (10,0.5){};
  \node[endo,label=left:{OutageIncident}]             (OI)  at (11,1.5)  {};

  %--------------- NÓ LATENTE ---------------
  \node[exo,label=left:{HeavyTraffic}] (HT) at (3,2.5) {};

  % ----------- ARESTAS SÓLIDAS (relações causais) -------
  \foreach \x/\y in {DBC/DBL, DBL/SBL, SBL/SAL, SBL/SBE,
                     SAL/SAT, SAT/SAC, SAC/OI,
                     SBE/SAE, SAE/OI}{
      \draw[->,>=latex] (\x) -- (\y);
  }

  % ----------- ARESTAS TRACEJADAS (confounding) ---------
  \foreach \x/\y in {HT/SBL, HT/SAL}{
      \draw[->,>=latex,dashed] (\x) -- (\y);
  }
\end{tikzpicture}
\vspace*{-2ex}
\caption{Model 2: Medium-scale microservice observability causal model. Dashed arrows indicate latent heavy traffic confounding latency between Microservice B and in Microservice A. Target node is the OutageIncident, indicating unavailability.}
\label{figure:db-latency-dag}
\end{figure}

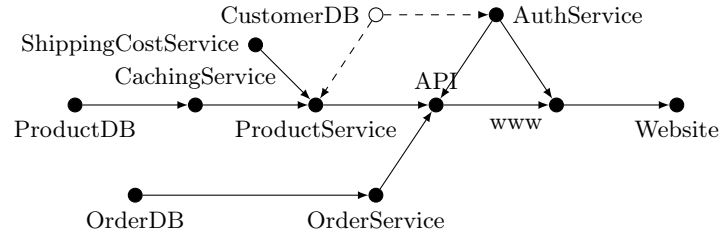
\begin{figure}[t]
\centering
\begin{tikzpicture}[scale=0.8]
  % ----------- estilos ------------------------------------
  \tikzstyle{endo} = [draw,circle,fill=black, inner sep=0.2em]
  \tikzstyle{exo}  = [draw,circle,fill=white,inner sep=0.2em]

  % ---------------------- NÓS ------------------------------
  % bancos (esquerda)
  \node[exo, label=left:{CustomerDB}] (CDB) at (7, 1.5)  {};
  \node[endo,label=below:{ProductDB}]  (PDB) at (2, 0)  {};
  \node[endo,label=below:{OrderDB}]    (ODB) at (3,-1.5)  {};

  % serviços intermediários
  \node[endo,label=left:{ShippingCostService}] (SCS) at (5, 1) {};
  \node[endo,label=above:{CachingService}]      (CS)  at (4, 0) {};
  
  \node[endo,label=right:{AuthService}]         (AS)  at (9, 1.5) {};
  \node[endo,label=below:{ProductService}]      (PS)  at (6, 0) {};
  \node[endo,label=below:{OrderService}]        (OS)  at (7,-1.5) {};

  % camada de API / front-end
  \node[endo,label=above:{API}] (API) at (8, 0) {};
  \node[endo,label=below left:{www}] (WWW) at (10,0) {};
  \node[endo,label=below:{Website}] (WEB) at (12,0) {};

  % ---------------- ARESTAS SÓLIDAS ------------------------
  \foreach \x/\y in {PDB/CS, CS/PS, SCS/PS,
                     PS/API, AS/API, OS/API,
                     API/WWW, AS/WWW,
                     WWW/WEB,
                     ODB/OS}
    \draw[->,>=latex] (\x) -- (\y);

  % --------------- ARESTAS TRACEJADAS (confusão) -----------
  \foreach \x/\y in {CDB/AS, CDB/PS}
    \draw[->,>=latex,dashed] (\x) -- (\y);

\end{tikzpicture}
\vspace*{-2ex}
\caption{Model 3: Finding the Root Cause of large latencies in a microservice architecture. The latency of  customers databases is not  monitored, so it is a latent confounder. Website is the target node for which we want to run a RCA.}
\label{figure:webapp-dag}
\end{figure}

Our results with experiments in Models 1-3 are summarized in Tables 1-2. For each model-narrative pair with   ground truth, we present the identified root cause when using each of the proposed metrics within the Algorithm~\ref{alg:dfs-short}. We can see that metrics derived from \textit{necessity} (PN and w-PN) consistently identifies the correct root cause pathways across various scenarios. In contrast, \textit{sufficiency}-based metrics (PS and w-PS) consistently underperformed, often leading to incorrect conclusions. PNS exhibited inconsistent performance across  causal narratives.

To  interpret our experimental findings, it is essential to clarify the specific nature of the RCA we are interested in. The umbrella term \textit{Root Cause Analysis} encompasses diverse techniques, ranging from analyses of isolated incidents, anomalies, or outliers occurring at specific points in time, to more generalized evaluations over extended periods, such as structural changes or system degradation. Even though counterfactual inference   emphasizes individual-level \textit{what-if} questions, our proposed method diverges slightly by aggregating binary observations over specified intervals, aiming to determine the general root cause behind persistent inconsistencies observed in a target variable.

%Another critical consideration pertains to the philosophical and practical dimensions of \textit{necessity} and \textit{sufficiency}. The concepts inherently implies temporal proximity: the closer a candidate cause is to the observed event, the stronger its contributory role appears due to intrinsic temporal dependencies in chain-reaction system failures. Conversely, necessity intuitively captures the idea that factors situated earlier in a causal chain must occur consistently to enable subsequent events. This temporal perspective underpins our analyses implicitly, even though explicit temporal variables were not directly modeled.

\begin{table}[t]
\centering
\caption{Causal Path-Ranking results with \textit{strong} probabilities $(\alpha = 2; w = 2)$}
\label{tab:exact-ranking}
\begin{tabular}{@{}c|l|l|l|l|l@{}} \toprule
\textbf{Model} & \textbf{Narrative} & \textbf{Ground Truth} &
\textbf{PN} & \textbf{PS} & \textbf{PNS} \\ \midrule
\multirow{4}{*}{M1}
 & N1 & MemLeak        & MemLeak        & MemLeak        & MemLeak       \\
 & N2 & HeavyTraffic   & ServiceCrash     & MemUsageHigh   & ServiceCrash   \\
 & N3 & ServiceCrash   & ServiceCrash   & ServiceCrash   & ServiceCrash   \\
 & N4 & MemUsageHigh   & MemUsageHigh   & MemUsageHigh   & MemUsageHigh   \\ \cmidrule{2-6}
\multirow{4}{*}{M2}
 & N1 & DB\_Latency    & DB\_Latency    & DB\_Latency    & DB\_Latency    \\
 & N2 & MS-A\_Threads  & MS-A\_Threads  & MS-A\_Threads  & MS-A\_Threads  \\
 & N3 & MS-B\_Error    & MS-B\_Error    & MS-B\_Error    & MS-B\_Error    \\
 & N4 & MS-A\_Latency  & MS-A\_Latency  & MS-A\_Threads  & MS-A\_Latency  \\ \cmidrule{2-6}
\multirow{2}{*}{M3}
 & N1 & CachingService & CachingService & ProductService & CachingService \\
 & N2 & OrderDB        & OrderDB        & API        & OrderService        \\ \bottomrule
\end{tabular}
\end{table}

\begin{table}[t]
\centering
\caption{Causal Path Ranking results with \textit{weak} measurements $(\alpha = 2; w = 2)$}
\label{tab:weak-ranking}
\begin{tabular}{@{}c|l|l|l|l@{}} \toprule
\textbf{Model} & \textbf{Narrative} & \textbf{Ground Truth} &
\textbf{weak PS} & \textbf{weak PN} \\ \midrule
\multirow{4}{*}{M1}
 & N1 & MemLeak        & NewDeploy        & MemLeak        \\
 & N2 & HeavyTraffic   & NewDeploy   & NewDeploy     \\
 & N3 & ServiceCrash   & NewDeploy   & ServiceCrash   \\
 & N4 & MemUsageHigh   & NewDeploy   & MemUsageHigh   \\ \cmidrule{2-5}
\multirow{4}{*}{M2}
 & N1 & DB\_Latency    & DB\_Change    & DB\_Change    \\
 & N2 & MS-A\_Threads  & MS-A\_Error  & MS-A\_Threads  \\
 & N3 & MS-B\_Error    & MS-A\_Crash    & MS-B\_Error    \\
 & N4 & MS-A\_Latency  & MS-A\_Threads  & MS-B\_Latency  \\ \cmidrule{2-5}
\multirow{2}{*}{M3}
 & N1 & CachingService & ProdutctService & CachingService \\
 & N2 & OrderDB        & API        & OrderService        \\ \bottomrule
\end{tabular}
\end{table}

Taking into account this qualitative assessment, we interpret the effectiveness of necessity metrics in RCA as inherently linked to their alignment with temporal causality,   distinguishing between preventive and remedial analyzes. If the goal is to identify critical components that impact system incidents, sufficiency measures may initially appear more informative. However, to  address underlying conditions triggering successive failures, necessity measures provide a more robust analytical foundation.

Accordingly, the consistent capability of necessity-based metrics to accurately recover the true causal paths underscores their adequacy for automated RCA in intelligent autonomous systems. From a practical perspective, the choice between PN and w-PN must consider computational and data requirements. While PN provides stronger discriminatory power, its inference involves substantially higher computational costs compared to the intervention-based w-PN. Thus, the decision to deploy either metric within intelligent RCA systems must carefully weigh their respective performance benefits against these operational constraints.

\section{Conclusion}\label{section:Conclusion}

To recap, we have presented contributions both on algorithmic and methodological aspects of causal reasoning. 

On algorithms, we presented simplifications to counterfactual graphs that lead to multilinear programs of reduced degree, by exploiting the information in input distributions. We have also presented a few additional simplifications to the computation of probabilities of causation PN, PS and PNS, as these counterfactual quantities are particularly important in practice. 

On methodological contributions for root cause analysis (RCA), we proposed a novel approach leveraging probabilities of causation as metrics for identifying and ranking causal paths within scenarios of limited observability, such as those in microservices observability contexts. 
By formulating RCA as the task of quantifying causal contributions of variables along potential causal paths, we developed and evaluated a ranking algorithm based on the concept of \textit{path significance}, where significance scores were systematically derived from  probabilities of causation (PN, weak-PN, PS, weak-PS, and PNS).
We also analyzed realistic scenarios with cascading failures and latent confounders, so as to understand the behavior of probabilities of causation in practice. Necessity-based measures PN and w-PN emerged as winners in identifying root causes and causal paths leading to observed incidents.

%This superior performance underscores the practical utility and theoretical soundness of necessity-based measures for RCA in systems characterized by chain-like dependencies, where upstream causes propagate through multiple service layers.

%In conclusion, our contributions facilitate both computational simplifications in counterfactual inference and robust methodological strategies in causation-based RCA, significantly enhancing the potential for automated, reliable root cause identification in complex software systems under limited observability. 
Looking at potential avenues for future research, we suggest extending our methods to recently formalized definitions of probabilities of causation for discrete and continuous variables. Such advances would  avoid  the need for data binarization and potentially enhance RCA accuracy and applicability.

\section*{Acknowledgements}

We thank the ICTi, Instituto de Ciência e Tecnologia Itaú, for providing key funding
for this work through the C2D - Centro de Ciência de Dados at Universidade de São Paulo.
F.G.C.\ was partially supported by CNPq grants 312180/2018-7 and 305753/2022-3. 
D.D.M.\ was partially supported by CNPq grant 305136/ 2022-4 and São Paulo Research Agency (FAPESP) grant 2022/02937-9.
The authors also thank support by CAPES - Finance Code 001.

Any opinions, findings, conclusions or recommendations expressed in this material are those of the authors and do not necessarily reflect the views of Itaú Unibanco and Instituto de Ciência e Tecnologia Itaú. All data used in this study comply with the Brazilian General Data Protection Law.

% ---- Bibliography ----
\bibliographystyle{splncs04}
\bibliography{rca} 

\begin{thebibliography}{10}
\providecommand{\url}[1]{\texttt{#1}}
\providecommand{\urlprefix}{URL }
\providecommand{\doi}[1]{https://doi.org/#1}

\bibitem{balke-pearl}
Balke, A., Pearl, J.: Counterfactual probabilities: Computational methods, bounds and applications. In: Conf.\ on Uncertainty in Artificial Intelligence. pp. 46--54 (1994)

\bibitem{Budhathoki2022}
Budhathoki, K., Minorics, L., Bl{\"{o}}baum, P., Janzing, D.: Causal structure-based root cause analysis of outliers. In: Int.\ Conf.\ on Machine Learning (ICML) (2022)

\bibitem{Ikram2022}
Ikram, A., Chakraborty, S., Mitra, S., Saini, S.K., Bagchi, S., Kocaoglu, M.: Root cause analysis of failures in microservices through causal discovery. In: Advances in Neural Information Processing Systems (NeurIPS) (2022)

\bibitem{Li2022}
Li, M., Li, Z., Yin, K., Nie, X., Zhang, W., Sui, K., Pei, D.: Causal inference-based root cause analysis for online service systems with intervention recognition. In: Proceedings of the 28th ACM SIGKDD Conference on Knowledge Discovery and Data Mining (KDD). pp. 1575--1585 (2022)

\bibitem{Mokhtari2023}
Mokhtari, A., Pang, H., Farakos, S.S., McKenna, C., Crowley, C., Cranford, V., Bowen, A., Phillips, S., Madad, A., Obenhuber, D., Doren, J.M.V.: Leveraging risk assessment for foodborne outbreak investigations: The quantitative risk assessment-epidemic curve prediction model. Risk Analysis  \textbf{43}(2),  324--338 (2023)

\bibitem{Oliveira2022}
Oliveira, E.E., Miguéis, V.L., Borges, J.L.: Understanding overlap in automatic root cause analysis in manufacturing using causal inference. IEEE Access  \textbf{10},  191--201 (2022). \doi{10.1109/ACCESS.2021.3139199}

\bibitem{causality}
Pearl, J.: Causality. Cambridge University Press (2009)

\bibitem{primer}
Pearl, J., Glymour, M., Jewell, N.P.: Causal inference in statistics: a primer. Wiley, Chichester, West Sussex (2016)

\bibitem{Richens2020}
Richens, J.G., Lee, C.M., Johri, S.: Improving the accuracy of medical diagnosis with causal machine learning. Nature Comm.  \textbf{11}, ~3923 (2020)

\bibitem{Shpitser2007}
Shpitser, I., Pearl, J.: What counterfactuals can be tested. In: Conference on Uncertainty in Artificial Intelligence. p. 352–359. AUAI Press (2007)

\bibitem{shridharan23icml}
Shridharan, M., Iyengar, G.: Causal bounds in quasi-{M}arkovian graphs. In: Proceedings of the 40th International Conference on Machine Learning. vol.~202, pp. 31675--31692. PMLR (2023)

\bibitem{Tian2002}
Tian, J.: Studies in Causal Reasoning and Learning. Ph.D. thesis, UCLA (2002)

\bibitem{Wang2024}
Wang, T., Qi, G.: A comprehensive survey on root cause analysis in (micro) services: Methodologies, challenges, and trends. arXiv preprint arXiv:2408.00803  (2024)

\bibitem{zaffalon2024}
Zaffalon, M., Antonucci, A., Caba\~nas, R., Huber, D., Azzimonti, D.: Efficient computation of counterfactual bounds. International Journal of Approximate Reasoning pp. 1--24 (2024)

\bibitem{zhang-tiam-bareinboim}
Zhang, J., Tian, J., Bareinboim, E.: Partial counterfactual identification from observational and experimental data. In: Int.\ Conf.\ on Machine Learning. vol.~162, pp. 26548--26558. PMLR (2022)

\end{thebibliography}

\end{document}